# A New Efficient Numbering System
## Application to Numbers Generation and Visual Markers Design

Messaoud Mostefai[1,2], Salah Khodja[1] and Youssef Chahir[2]

[1] Ultimara Company, 370 Convention Way, Redwood City, CA 94063, USA,
[2] Computer Science Department, University of Caen, 14032 Caen, France

Corresponding author
Email address: youssef.chahir@unicaen.fr (Youssef Chahir)

**Abstract**

This short paper introduces a recently patented line based numbering system. The last allows a best concordance with decimal digits values, and open up new opportunities, which are not possible with the classical decimal numeration system. Proposed OILU symbolic allows generating a new type of number series, based on multi facets numbers splitting process. On the other hand, this new symbolic is used in the development of new visual markers, highly required in augmented reality and UAV's navigation applications.

## 1. Background

Numbers are by far the most important thing human beings have ever invented to manage their daily lives, starting from basic counting to the resolution of complex problems [1, 2]. Although they have succeeded in computerizing their calculation and coding faculties, humans continue to use a symbol-based numbering system that is not well suited to their machines. Fortunately, this has not prevented the two parts from communicating and exchanging information through more or less complex interfaces, like Bar and QR codes readers [3,4]. Nevertheless, rapprochement efforts are often on the machine side because obviously, the human being is by nature hostile to habits changes, especially if these habits are ancestral.

In this work, a new human-machine readable numeration system is proposed [5]. The last allows efficient and less computational numbers recognition and exploitation. The line-based nature of OILU symbols as well as their pyramidal disposition, allows manipulating numbers as multi-facets objects and generating highly distinguishable real time fiducials. Applications are diverse and varied, starting from authentication to visual human-machine communication protocols. However, we will limit ourselves to the presentation of the OILU system and its application to visual data representation and codification.

The remainder of the paper is structured as follows: Section 2 details the adopted approach for the construction of the OILU numeration system. Section 3 presents the developed number series generator. Section 4 shows the development of real time visual OILU markers and briefly describe their advantages. Finally, primarily conclusions are drawn in Section 5.

## 2. OILU Numeration System Development

An interesting book written by Georges Ifrah in 1985, entitled "From One to Zero" [2], illustrates the influence that the surrounding environment has had on human beings in their choice of a numbering systems. In the same context, folding a segment in two, then three, and finally four sides square (figure 1.a), allows generating group of basic patterns that can be used as a numbering system. The latter is composed of ten symbols, generated as follows: the first four symbols { |, ⌐, ⊔ and ☐ }, are respectively affected to digits one, two, three and zero. The six lacking symbols are then produced by successive rotations (quarter counterclockwise) of the two symbols {⌐ and ⊔}, (FIG.1.b). Rotating {⌐} symbols are affected to even digits, and rotating {⊔} symbols are affected to odd digits.

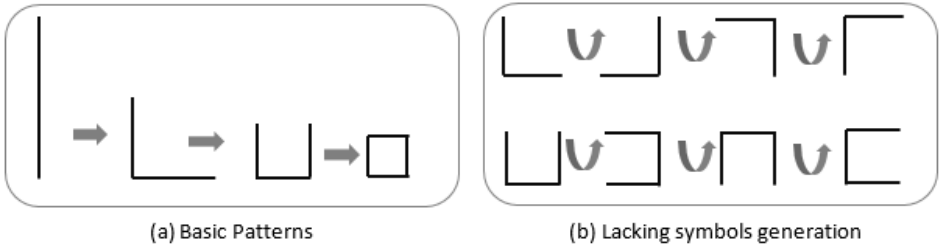

**Fig.1.** Symbols generation

If we exclude { | and ☐ } symbols which are not concerned by the rotation and are automatically affected to digits «0» and «1», the rest of symbols are affected (according to their orientations) to the corresponding 8 directions (Figure 2.a). The following table (Figure 2.b) presents the correspondence between the generated symbols and the decimal digits values. Energically speaking, OILU symbolic allows a reduction of at least 50% of the display energy: 25 segments versus 49 (for classical displayers) (FIG.3a). But the main interesting thing with this system, is that it allows

superimposing symbols in pyramidal form without losing the value of constructed numbers. Figure.3.b presents an example of a pyramidal representation of the decimal number (3172). The lecture sense is from the outside to the inside. Unlike the classical numeral decimal symbolic, the OILU symbolic allows to see a number as an object with its four facets. Indeed, we can extract from each point of view a different number value. Thus, a group of related numbers is formed. In case of our example, related numbers are : 3172 – 9158 – 7136 – 5194. Any facet's value allows deducing other facet's values.

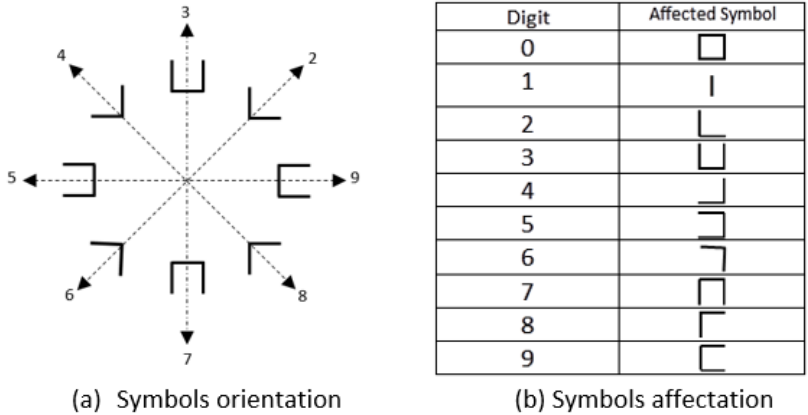

**FIG.2.** OILU Symbols affectation and codification

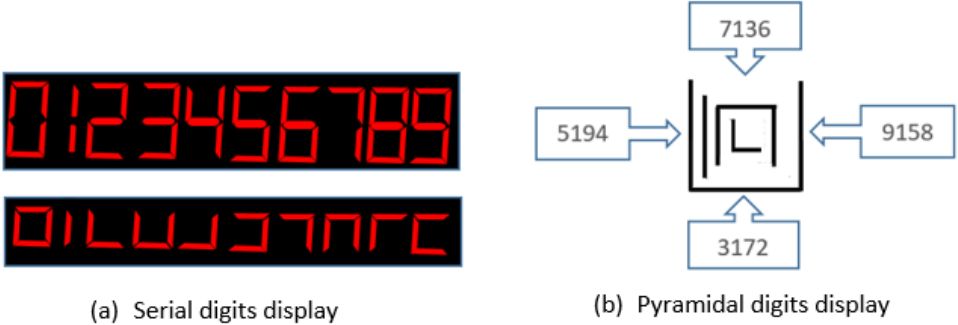

**FIG.3.** OILU Digits and Numbers display

In the following, we will present an example of a mathematical OILU application, which consist on generating number series from the well-known decimal and hexadecimal numeration systems. Such application confirms well that OILU symbolic is visually a set of Decimal and Hexadecimal numeration sets. Apart from this, the application itself can be the basis of efficient data encryption and rekeying schemes [6,7].

## 3. A New Number Series Generator

If we look at the seven-segment digits of any digital clock (Fig.4a), we will see that the last are simply composed of OILU symbols. This interesting property allows splitting any seven segments digit into two OILU symbols, according to one of the three splitting strategies (a), (b) or (c) (FIG.4b). The first strategy (a) consider the central horizontal segment as a shareable segment between the two parts. The second strategy (b) consider the central horizontal segment as belonging to the upper part, and finally the last strategy (c) consider the central segment as belonging to the lower part. FIG.5 presents the Dec/Hex digits splitting into OILU symbols, starting from the digit "0" to digit "F". To ensure a total reversibility, and in order to avoid any confusion, produced extra symbols (not OILU symbols) are replaced with symbols produced by the splitting strategy (a), which is free of any extra symbol.

Such decomposition allows generating a new type of number series based on an alternative transition between Decimal/Hexadecimal (Dec/Hex) and OILU symbols. Fig.6, shows examples of digit and number splitting using different strategies. Depending on the starting point and the adopted splitting strategy, a list of linked numbers is produced. To enlarge the amount of generated numbers, successive splitting operations can be performed starting from the different numbers facets. Thus, various number series are produced according to pre-established navigation rules (chosen facet and splitting strategy).

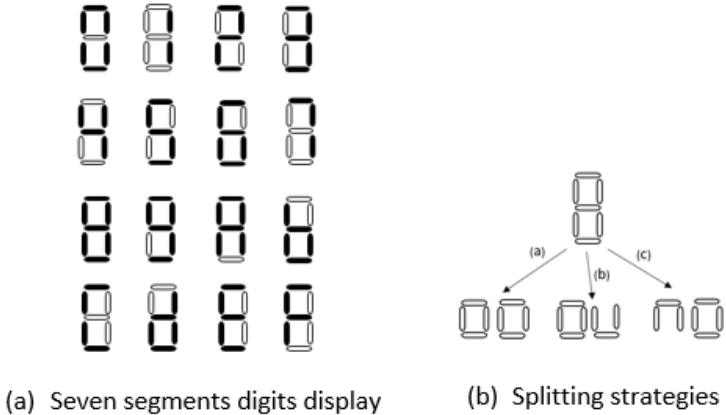

(a) Seven segments digits display    (b) Splitting strategies

**FIG.4.** Seven segments digits splitting strategies

The following application is another example of a new visual data representation, widely used in Augmented Reality and UAV's navigation applications.

## 4. New Real Time OILU Markers

Markers are widely used in Augmented Reality and UAV's navigation applications [8,9]. They provide additional information to computer vision software for efficient localization and camera pose estimation. In our case, developed OILU symbolic allowed us to generate real time visual makers composed of a group of superimposed OILU symbols (FIG.7.a). Based on easy data coding scheme, developed markers enable producing a large panel of unique real time identifiers with highly distinguishable patterns. Compared to the state of art 2D markers [10], OILU markers does not require any geometrical transformation for their identification (FIG.7b), allowing thus a simpler and less computational processing. Developed identification method as well as real experimental tests will be presented in a dedicated article.

## 5. Preliminary Conclusion

The main objective of this short paper is to introduce this new numeration system to the scientific community. The proposed symbolic allows to view numbers differently, opening by the way new opportunities which are not possible with the classical numeration system. The most important features are symbols superimposition, multi facets numbers and numbers splitting.

Presented Number Series Generator is an example of several OILU applications that are under development. Also, presented fiducials, are an example of new visual data representation and codification, widely used in Augmented Reality and UAV's navigation applications. Based on easy data coding scheme, developed fiducials enable producing a large panel of unique real time identifiers with highly distinguishable patterns.

OILU symbolic is in its early stage. Collaborative works are engaged with several researchers in order to show the effectiveness of this new symbolic in various engineering fields.

For tests, a dedicated web site (http://oilucode.net) is under development to allow downloading our OILU images Database. More details and applications will follow.

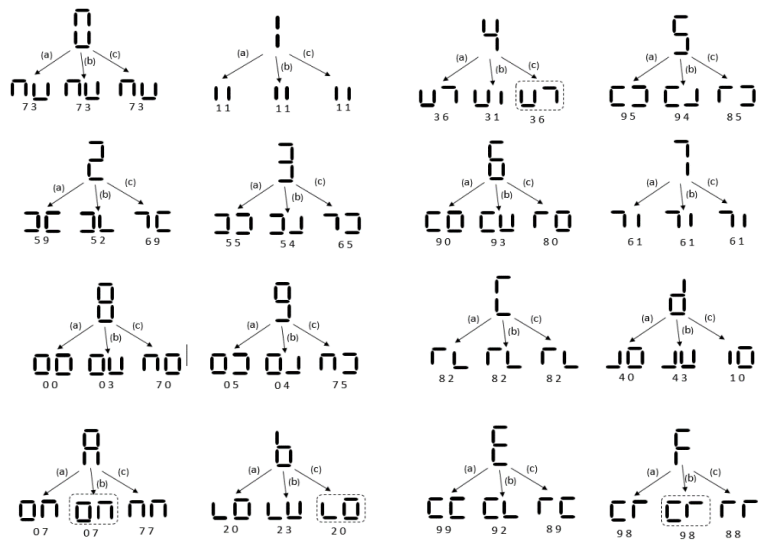

**FIG.5.** Decimal/Hexadecimal Digits Splitting

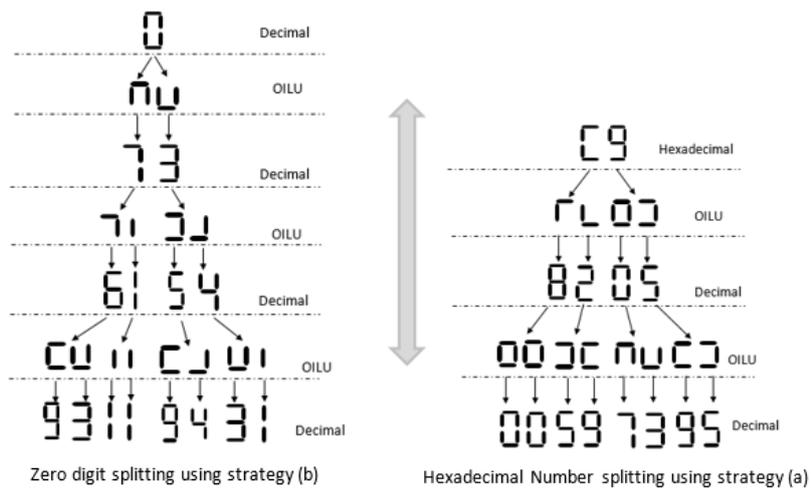

**FIG.6.** Examples of digit and number splitting results

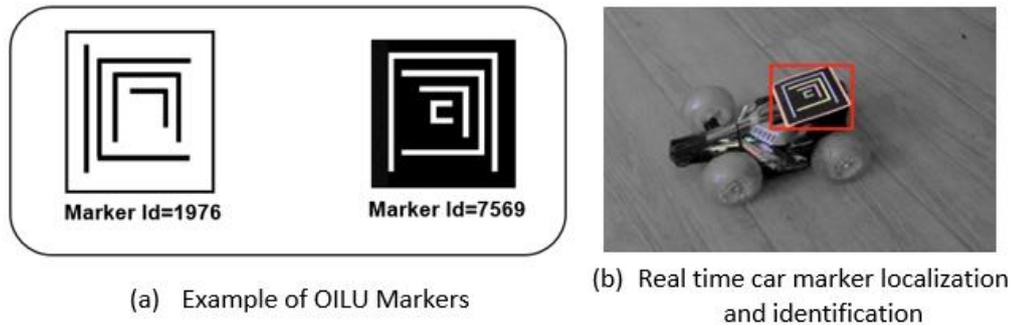

**FIG.7.** Real time OILU Markers